\documentclass[11pt]{article}

\usepackage[final]{acl}
\usepackage{times}
\usepackage{latexsym}
\usepackage[T1]{fontenc}
\usepackage[utf8]{inputenc}
\usepackage{microtype}
\usepackage{inconsolata}
\usepackage{graphicx}
\usepackage{booktabs}
\usepackage{amsmath}
\usepackage{amssymb}
\usepackage{multirow}
\usepackage{url}
\usepackage{makecell}
\usepackage{tikz}
\usetikzlibrary{positioning,calc,arrows.meta,fit,backgrounds}
\usepackage{float}
\usepackage{tabularray}

\UseTblrLibrary{booktabs}
\usepackage[table]{xcolor}

\definecolor{groupB}{HTML}{ECEBE5}
\definecolor{grpProg}{HTML}{E7EAFA}
\definecolor{grpResearch}{HTML}{E8EEEA}
\definecolor{grpEnt}{HTML}{F4E8DF}
\definecolor{grpGui}{HTML}{EDE6EF}
\definecolor{grpFC}{HTML}{F3E5E5}
\definecolor{grpP}{HTML}{E7EAFA}
\definecolor{messierBlue}{HTML}{3A55D6}

\usepackage{pifont}
\newcommand{\cmark}{\ding{51}}
\newcommand{\xmark}{\ding{55}}

\usepackage{xspace}
\newcommand{\messier}{\textsc{Messier}\xspace}

\title{Messier: A High-Resolution Corpus for Cross-Benchmark Agent Evaluation}

\author{
  Stefan Krsteski, Charlotte Meyer, Guillaume Allegre, \\
  \textbf{Tony O'Halloran}, \textbf{Alexandre Sallinen} \\
  Andromede AI \\
  \small \textbf{Correspondence:} \href{mailto:stefan@andromede.ai}{stefan@andromede.ai}
}
\begin{document}
\maketitle

\begin{abstract}

Comprehensively evaluating AI agents across interactive environments is difficult due to fragmented tasks, scaffolds, verifiers, and scoring rules. Unfortunately, existing efforts to unify these evaluations are limited in scale and domain, making costly reruns necessary and leaving available data incomparable. We introduce \messier, a unified corpus of 923{,}158 records spanning 30 benchmarks, 745 agents, 11{,}891 tasks, and 71{,}969 verifiers. \messier combines public evaluation results with new runs on six underrepresented professional and scientific benchmarks, standardizing their heterogeneous components into a common schema. Using this corpus, we show that frontier progress is uneven across benchmark groups, with function-calling evaluations largely saturated, programming improving fastest, and enterprise workflows remaining most challenging. Counterfactual rescoring further shows that reported performance can be highly sensitive to how verifier results are aggregated. Finally, we derive capability scores from our corpus that correlate with Epoch's Evaluation Capability Index rankings at Spearman $\rho = 0.84$. The scores can also be estimated for subsets defined by domain, occupation, action space, or verifier type. In essence, \messier is a reusable resource for studying agent performance at scale, and a basis for designing better evaluations.

\end{abstract}
\begin{table*}[!htbp]
  \centering
  \small
  \setlength{\tabcolsep}{6pt}
  \renewcommand{\arraystretch}{1.20}

\begin{tabular}{l c r r r r r}
\toprule
    \textbf{Resource} & \textbf{Agentic} & \textbf{Benchmarks} & \textbf{Environments} & \textbf{Models} & \textbf{Scaffolds} & \textbf{Tasks} \\
    \midrule
    HELM~\cite{liang2022holistic}                            & \xmark & 42  & N/A     & 30   & N/A & 30{,}000 \\
    METR~\cite{kwa2026measuring}                              & \cmark & 3   & 3       & 13   & N/A & 170      \\
    Epoch ECI~\cite{ho2025rosetta}                       & \cmark & 37  & N/A     & 200  & N/A & N/A      \\
    TOUCAN~\cite{xu2025toucan}                         & \cmark & N/A & 495     & 3    & 2 & 1.5M     \\
    HAL~\cite{kapoor2025holistic}                             & \cmark & 9   & N/A     & 9    & $\sim$10 & $\sim$1{,}500 \\
    BRIDGE~\cite{liu2026bridge}                          & \cmark & 4   & N/A     & 176  & $\sim$70 & 752      \\
    Agent Psych.~\cite{ge2026agent}    & \cmark & 4   & N/A     & 275  & $\sim$90 & 1{,}421  \\
    \midrule
    \cellcolor{grpP}\textbf{\messier}
      & \cellcolor{grpP}\cmark
      & \cellcolor{grpP}30
      & \cellcolor{grpP}2{,}495
      & \cellcolor{grpP}343
      & \cellcolor{grpP}221
      & \cellcolor{grpP}11{,}891 \\
    \bottomrule
  \end{tabular}
  \caption{\textbf{Comparison with related collections of LLM evaluation data.} Counts are as reported by each resource at release. $\sim$ denotes approximate values inferred from the corresponding paper. For \messier, environments are counted by distinct normalized environment identifier. N/A means the information is not reported or applicable.}
  \label{tab:comparison}
\end{table*}

\section{Introduction}
\label{sec:intro}

Agent benchmarks are a central instrument for measuring how language models act in interactive environments. They test whether these systems can complete professional work under domain-specific constraints, with the evidence and tools they are given. Their outputs can influence scientific claims \cite{bragg2026astabench}, clinical records \cite{jiang2025medagentbench}, legal matters \cite{harvey2026lab}, software \cite{jimenez2024swe}, and financial decisions \cite{bigeard2025finance}.

Their proliferation across domains has made results difficult to compare. Benchmarks vary in domain, agent definition, environment, action space, and success condition~\cite{kapoor2025holistic}. Final scores compound the problem by compressing complex behaviors into a single number. A math task may use one binary check, whereas a multi-step business workflow may combine several checks into a final result. In both cases, success need not represent the same accomplishment~\cite{burnell2023rethink,yang2026benchmark,sha2026benchscope}.

To make heterogeneous results comparable, newer frameworks offer unified measurements. For instance, Epoch's Evaluation Capability Index (ECI)~\cite{ho2025rosetta} applies Item Response Theory (IRT)~\cite{rasch1960probabilistic} to map diverse benchmark scores onto a shared scale of model capability and task difficulty. A complementary line of work, METR~\cite{kwa2026measuring}, measures agent capacity by task horizon length, showing that model execution limits have doubled roughly every seven months since 2019.

The data underlying these scores remain incomplete. Epoch does not release results for individual tasks, but only aggregates per benchmark, and focuses primarily on coding and math. In contrast, METR provides granular per-task data, but for only 170 tasks~\cite{kwa2026measuring}. Consequently, extending these analyses often requires reconstructing the evaluation corpora from scratch and rerunning costly agent rollouts, with up-to-date large-scale sweeps costing roughly \$40{,}000~\cite{kapoor2025holistic}. Existing meta-collections partly address this gap, but target narrow agentic settings, cover only static benchmarks, or remain small~\cite{liang2022holistic,srivastava2023beyond,lu2025agentrewardbench,xu2025toucan}. In parallel, the emergence of reward hacking~\cite{thaman2026reward,atinafu2026rewardhackingagents,gabor2025evilgenie}, evaluation awareness~\cite{needham2025large,hua2025steering}, and sandbagging~\cite{van2025ai} further highlights the need for open data.

We introduce \messier, a corpus of 923\,k records combining public results from 24 benchmarks with evaluations on six more, spanning quantum algorithm design, legal and clinical workflows, and other under-evaluated professional domains. With SOC and NAICS task classifications~\cite{bls2018soc,omb2022naics} and fine-grained records of agents, tasks, and verifiers, \messier supports analyses across industries, as well as studies of agent behavior, verifier design, and evaluation failure modes. Our main contributions are:

\begin{itemize}
    \item \textbf{A unified agent evaluation corpus} consolidating 30 benchmarks (with new runs on six), 745 agents, 11{,}891 tasks, and 71{,}969 verifiers (\S\ref{sec:messier}), providing both macro- and micro-scale views of agent performance.
    
    \item \textbf{Capability indices from open data} that correspond to Epoch ECI rankings at Spearman $\rho = 0.84$ (\S\ref{sec:eci-reproduction}) and support slicing by occupation, action space, or verifier type.

    \item \textbf{Evidence that scoring rules can change measured performance}, obtained by rescoring stored verifier results counterfactually. For example, all-pass scoring can report near-zero task success even when many verifier results are positive.
\end{itemize}
By enabling researchers to reuse, share, and contribute evaluation results through a single corpus, \messier aims to reduce redundant compute.\footnote{Code and data: \href{https://github.com/Andromede-AI/messier}{github.com/Andromede-AI/messier}}

\section{Related Work}
\label{sec:related}

\paragraph{Agent benchmarks.}
Evaluating models within interactive environments shifts the testing paradigm away from static text pairs (e.g. QA) toward sequential execution, where an agent receives observations over multiple turns, selects actions, modifies files, or invokes tools before a verifier produces a result. Recent benchmarks implement this setup across several environment types, including software engineering~\cite{jimenez2024swe,shetty2026gso,jain2025livecodebench}, web navigation~\cite{xue2025illusion,wei2025browsecomp,he2024webvoyager}, terminal and operating-system tasks~\cite{merrill2026terminal,xie2024osworld}, tool-use and function-calling environments~\cite{patil2025berkeley,barres2025tau,wang2025mcp}, security tasks~\cite{zhang2025cybench}, economic work~\cite{patwardhan2025gdpval,xu2026theagentcompany}, and professional domains such as science, medicine, and law~\cite{harvey2026lab,bragg2026astabench,jiang2025medagentbench}. Each benchmark, however, defines its own naming conventions, scaffolding interfaces, and underlying verifier architectures. Consequently, cross-benchmark comparison requires significant manual effort and currently relies on post-hoc consolidation or expensive reruns. To overcome this bottleneck, \messier acts as an open aggregator across these fragmented domains, sparing researchers the prohibitive and redundant compute costs.

\paragraph{Evaluation collections.}
Parallel efforts increasingly focus on consolidation by assembling heterogeneous execution traces into unified formats. These efforts range from non-agentic text collections such as HELM~\cite{liang2022holistic} and BIG-bench~\cite{srivastava2023beyond} to collections built around interactive environments~\cite{liu2024agentbench, mialon2024gaia, ma2024agentboard}. However, current repositories remain limited by their specific operational scopes or focus areas. For instance, AgentRewardBench~\cite{lu2025agentrewardbench} focuses on evaluating trajectory judges, while TOUCAN~\cite{xu2025toucan} provides synthetic data optimized for agent training rather than for empirical capability indexing and analysis. The closest precedent to our work is the Holistic Agent Leaderboard (HAL)~\cite{kapoor2025holistic}, which introduces an active, cost-controlled evaluation harness and a corpus of over 21{,}000 agent rollouts across nine benchmarks. HAL addresses the problem of producing comparable new evaluations under a shared execution protocol. However, much of the empirical record of agent evaluation already exists, dispersed across public releases with incompatible namings, agent metadata, scoring semantics, verifier definitions, and trace formats. As a result, existing results remain difficult to integrate into broader comparative analyses.

\paragraph{Heterogeneity in benchmark evaluation.}
A growing body of literature shows that aggregate benchmark scores frequently conceal information visible only at the instance level, that is, at individual tasks or verifier results within a benchmark~\cite{schaeffer2023emergent, madaan2024quantifying}. BenchScope~\cite{sha2026benchscope} demonstrates that per-instance analysis reveals roughly five times as many latent dimensions in model performance as macro-level analysis, suggesting that aggregate benchmark scores conflate multiple distinct capabilities. Similarly, \citet{yang2026benchmark} document that language models achieving identical aggregate accuracy can still disagree on 16\% to 66\% of individual items, creating risks for conclusions drawn purely from macro totals. This risk extends to automated evaluation layers: CARE~\cite{zhao2026care} shows that aggregation under correlated judge errors can distort leaderboard rankings, while SkillsBench~\cite{li2026skillsbench} and ACE-Bench~\cite{chen2025acebench} show that final scores are sensitive to benchmark composition, including the mix of domains, skills, and task horizons. If aggregate scores collapse differences in tasks and verification rules into a single metric, then macro-level comparisons can misrepresent model capability. We address this with a cross-benchmark corpus that preserves task and verifier results and classifies each task by occupation to characterize variation in domain composition.

\begin{figure*}[t]
  \centering
  \resizebox{0.88\textwidth}{!}{%
  \begin{tikzpicture}[
    font=\small,
    >=Stealth,
    box/.style={draw=black!75, fill=white, rounded corners=2pt,
                align=center, minimum height=6.5mm, inner sep=3pt},
    input/.style={box, draw=black!50, fill=messierBlue!4,
                  minimum width=18mm},
    trial/.style={box, draw=black!50, fill=messierBlue!10,
                  minimum width=17mm},
    results/.style={box, draw=black!50, fill=messierBlue!18,
                    minimum width=30mm},
    result/.style={box, draw=black!50, fill=messierBlue!26,
                   minimum width=22mm},
    card/.style={box, minimum width=18mm, minimum height=5.5mm,
                 font=\scriptsize},
    taskcard/.style={card, draw=black!50, fill=messierBlue!12},
    verifiercard/.style={card, draw=black!50, fill=messierBlue!18},
    subgroup/.style={draw=black!25, fill=white, rounded corners=3pt,
                     inner sep=5pt},
    environment/.style={draw=black!35, fill=black!2,
                        rounded corners=4pt, inner sep=6pt},
    compose/.style={draw=black!65, thick},
    flow/.style={->, draw=black!80, thick},
    context/.style={->, draw=messierBlue!60, thick, dashed},
    note/.style={font=\scriptsize, fill=white, inner sep=1pt}
  ]
    \node[input] (agent) {Agent};
    \node[input, above left=1mm and 8mm of agent] (model) {Model};
    \node[input, below left=1mm and 8mm of agent] (scaffold) {Scaffold};

    \node[trial, right=10mm of agent] (trial2) {Trial $2$};
    \node[trial, above=4mm of trial2] (trial1) {Trial $1$};
    \node[trial, below=6mm of trial2] (trialJ) {Trial $J$};
    \node[font=\scriptsize, align=center, inner sep=0pt] (trialdots)
      at ($(trial2.south)!0.5!(trialJ.north)$)
      {$\cdot$\\[-3.5pt]$\cdot$\\[-3.5pt]$\cdot$};

    \node[results, right=7mm of trial1] (vr1)
      {Verifier results\\$r_{11},\ldots,r_{N1}$};
    \node[results, right=7mm of trial2] (vr2)
      {Verifier results\\$r_{12},\ldots,r_{N2}$};
    \node[results, right=7mm of trialJ] (vrJ)
      {Verifier results\\$r_{1J},\ldots,r_{NJ}$};

    \node[result, right=8mm of vr1] (res1) {Trial result $S_1$};
    \node[result, right=8mm of vr2] (res2) {Trial result $S_2$};
    \node[result, right=8mm of vrJ] (resJ) {Trial result $S_J$};

    \node[taskcard] (tK) at ($(trial1.north)+(3mm,10mm)$) {Task $K$};
    \node[taskcard] (t2) at ($(trial1.north)+(1.5mm,11.5mm)$) {Task $2$};
    \node[taskcard] (t1)
      at ($(trial1.north)+(0,13mm)$) {Task $1$};

    \node[verifiercard] (vN) at ($(vr1.north)+(3mm,10mm)$) {$v_N$};
    \node[verifiercard] (v2) at ($(vr1.north)+(1.5mm,11.5mm)$) {$v_2$};
    \node[verifiercard] (v1)
      at ($(vr1.north)+(0,13mm)$) {$v_1$};

    \node[fit=(t1)(tK), inner sep=5pt] (taskfit) {};
    \node[fit=(v1)(vN), inner sep=5pt] (verifierfit) {};
    \coordinate (envtop) at ($(taskfit.north)+(0,8mm)$);
    \begin{scope}[on background layer]
      \node[environment, fit=(taskfit)(verifierfit)(envtop)] (env) {};
      \node[subgroup, fit=(t1)(tK),
            label={[font=\scriptsize]above:Tasks}] (taskbox) {};
      \node[subgroup, fit=(v1)(vN),
            label={[font=\scriptsize]above:Verifiers for Task $1$}]
            (verifierbox) {};
    \end{scope}
    \node[font=\small, anchor=north]
      at ($(env.north)+(0,-2pt)$) {Environment};

    \draw[compose] (model.east) -- (agent.west);
    \draw[compose] (scaffold.east) -- (agent.west);

    \draw[flow] (agent.east) -- (trial1.west);
    \draw[flow] (agent.east) -- (trial2.west);
    \draw[flow] (agent.east) -- (trialJ.west);

    \draw[flow] (trial1.east) -- (vr1.west);
    \draw[flow] (trial2.east) -- (vr2.west);
    \draw[flow] (trialJ.east) -- (vrJ.west);

    \draw[flow] (vr1.east) -- node[above,note] {$f$} (res1.west);
    \draw[flow] (vr2.east) -- node[above,note] {$f$} (res2.west);
    \draw[flow] (vrJ.east) -- node[above,note] {$f$} (resJ.west);

    \node[fit=(trial1)(trial2)(trialJ), inner sep=1pt] (trialgroup) {};
    \node[fit=(vr1)(vr2)(vrJ), inner sep=1pt] (vrgroup) {};
    \draw[context] (t1.south) -- (trialgroup.north);
    \draw[context] (v1.south) -- (vrgroup.north);
  \end{tikzpicture}
  }
  \caption{\textbf{Evaluation flow.} An environment supports one or more tasks, each associated with one or more verifiers. A model and scaffold form an agent, which may execute a selected task over repeated trials $\tau_j$. Applying verifier $v_i$ to trial $\tau_j$ produces $r_{ij}=v_i(\tau_j)$. The scoring rule $f$ maps each trial's verifier results to its trial result $S_j$.}
  \label{fig:schema-er}
\end{figure*}
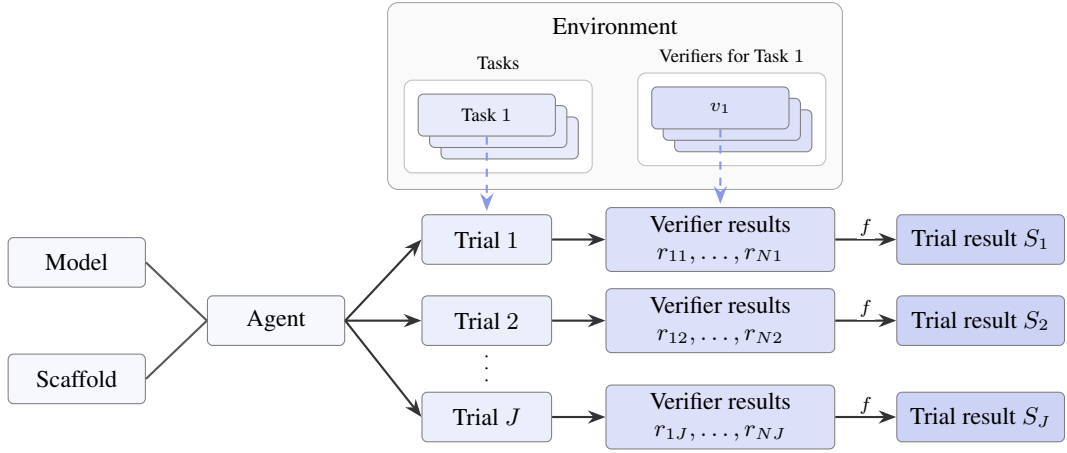

\section{Background}
\label{sec:preliminaries}

In an interactive evaluation, a large language model (LLM) acts through a scaffold, receives feedback from an environment, and produces a trial, which one or more verifiers evaluate. Across benchmarks, the boundaries between task, environment, and verifier are drawn inconsistently. We therefore define these components below, illustrate their relationships in Figure~\ref{fig:schema-er}, and summarize the terminology in Appendix Table~\ref{tab:terminology}.

\paragraph{Agents and environments.}
An \emph{agent} is a pair $(\textit{model}, \textit{scaffold})$: the model produces text, while the scaffold parses that text into actions, executes them, and returns observations. We treat the scaffold as part of the agent because the same model can yield substantially different benchmark results under different scaffolds~\cite{epoch2025whybenchmarkingishard, kapoor2025holistic}. 

An \emph{environment} is the system with which the agent interacts. This interaction is often formalized as a partially observable Markov decision process (POMDP)~\cite{kaelbling1998planning}. For our purposes, the relevant components are the state space $\mathcal{S}$, action space $\mathcal{A}$, and observation space $\mathcal{O}$, which define the possible states of the environment, the actions available to the agent, and the possible observations the agent can receive. In LLM evaluations, the state is typically the current configuration of an external system, such as a database or codebase, and observations are textual or visual renderings of that state, such as command outputs or web pages.

\paragraph{Tasks and verifiers.}
A \emph{task} \(T=(I,s_0)\) pairs an instruction \(I\) with an initial environment state \(s_0\in\mathcal S\). A \emph{trial} \(\tau\), also termed a \emph{rollout}, is one execution of an agent on a task. Each task is associated with $N$ \emph{verifiers} $\{v_1,\ldots,v_N\}$, typically programmatic checks, LLM judges, exact-match checks, or human evaluations. Applying verifier $v_i$ to trial $\tau$ produces a \emph{verifier result}
\[
    r_i = v_i(\tau),
\]
which may be binary or numerical depending on the benchmark.

A benchmark-defined \emph{scoring rule} $f$ maps the verifier results to the \emph{trial result} $S \in \{0,1\}$,
\[
    S = f(r_1,\ldots,r_N).
\]
We consider three scoring rules: direct, all-pass, and threshold. \emph{Direct} scoring uses a single binary verifier result as the trial result, so $S=r_1$. For tasks with multiple binary verifier results, \emph{all-pass} scoring assigns $S=1$ only if every verifier passes:
\[
    S = \prod_{i=1}^{N} r_i.
\]
A \emph{threshold} rule instead compares a benchmark-defined summary $g$ of the verifier results with a threshold $\gamma$:
\[
    S =
    \mathbb{I}\!\left(
        g(r_1,\ldots,r_N) \geq \gamma
    \right).
\]
The function \(g\) aggregates verifier results, for example by taking their mean, or acts as the identity for a single verifier that returns a continuous score.

\paragraph{A reconciliation problem.}
Within a single benchmark, these choices are typically fixed. Across benchmarks, however, they create a reconciliation problem: scaffolds must be normalized alongside model names, and scores must be interpreted in light of differences in verifiers and scoring rules. For multi-verifier tasks, the map \(f\) is generally many-to-one, so \(S\) alone does not recover the verifier results that produced it. We therefore preserve $\{r_i\}$ alongside $S$ whenever available. For unsegmented verifiers like monolithic scripts or LLM judges, this result is the finest detail consistently available.

\section{\messier Dataset}
\label{sec:messier}

To construct \messier, we combine existing public releases with new evaluations targeting benchmarks where per-task data are scarce or absent. The sources include prior capability-measurement releases (METR~\cite{kwa2026measuring}, BRIDGE~\cite{liu2026bridge}, Agent Psychometrics~\cite{ge2026agent}), General AgentBench~\cite{li2026benchmark}, public benchmarks released by individual benchmark authors, and runs targeting a common set of five agents that we collected through the Harbor harness \cite{Harbor_Framework}. Table~\ref{tab:comparison} summarizes the resulting corpus alongside related collections. The source releases are described next.

\paragraph{METR}~\cite{kwa2026measuring} measures agent capacity by task-horizon length, defined as the duration of tasks a model can reliably complete when calibrated against human completion times. METR released per-task results for three time-horizon benchmarks (HCAST, SWAA, RE-Bench~\cite{wijk2024re}) used in this measurement. We ingest these results directly.

\paragraph{BRIDGE}~\cite{liu2026bridge} extends METR's task-horizon framing with a 2PL item response theory model, fitting per-agent ability and per-task difficulty jointly across METR-band tasks. The BRIDGE release contributes new evaluations on MLE-bench~\cite{chan2025mle}, GDPval~\cite{patwardhan2025gdpval}, and CyBench~\cite{zhang2025cybench}, plus human-time annotations we use for SWE-bench Verified~\cite{jimenez2024swe}.

\paragraph{Agent Psychometrics}~\cite{ge2026agent} is a recent effort to formalize psychometric measurement for language-model agents, fitting IRT difficulty parameters across a focused set of agent benchmarks. We use their release for per-task results on TerminalBench~\cite{merrill2026terminal}, GSO~\cite{shetty2026gso}, SWE-bench~\cite{jimenez2024swe}, and SWE-bench Pro~\cite{deng2025swe}.

\paragraph{General AgentBench}~\cite{li2026benchmark} releases agent trajectories spanning search, coding, mathematical reasoning, and tool-orchestration benchmarks. We use this release for per-task results on MathHay~\cite{wang2024mathhay}, BrowseComp~\cite{wei2025browsecomp}, WebVoyager~\cite{he2024webvoyager}, and MCPBench~\cite{wang2025mcp}.

\paragraph{Contributed runs.}
Public data are unevenly available, as running trials is expensive at frontier-model scale, and several recent efforts release only aggregate leaderboard scores. 
To fill these gaps, we contribute 
new evaluations on six benchmarks: HarveyAI-Lab~\cite{harvey2026lab}, MedAgentBench~\cite{jiang2025medagentbench}, DABStep~\cite{egg2025dabstep}, QCircuitBench~\cite{yang2026qcircuitbench}, ScienceAgentBench~\cite{chen2025scienceagentbench}, and ReplicationBench~\cite{ye2025replicationbench}. We evaluate the same five models on each benchmark using the OpenHands~\cite{wang2024openhands} scaffold through Harbor. The total cost was approximately \$10{,}000. All new runs include full execution traces (Appendix~\ref{app:compute}).

\paragraph{Public benchmark releases.}
For the majority of benchmarks (HLE~\cite{phan2025humanity}, LiveCodeBench, GDPval, MathArena~\cite{balunovic2026matharena}, SWE-bench, BFCL, OSWorld, $\tau^2$-bench, and others), we obtain per-task results from author-released artifacts, either as Hugging Face datasets or directly from benchmark GitHub repositories. When needed, we supplement primary releases with auxiliary sources, such as Hugging Face mirrors and leaderboard CSV. Appendix~\ref{app:sources} lists full source URLs per benchmark.

After consolidation, the corpus contains 30 benchmarks, 2{,}495 environments, 11{,}891 tasks, 71{,}969 verifiers, and 923{,}158 records across 343 models and 221 scaffolds (745 distinct agents). The data are summarized in Table~\ref{tab:per-bench}.

\paragraph{Reconciliation pipeline.}
The pipeline consists of four stages. 
First, we cast each incoming record into a uniform data model derived from \S\ref{sec:preliminaries}. Every environment groups one or more tasks, and every task carries one or more verifiers (Python script, exact-match check, LLM judge, or human label) together with a scoring rule. Each task also includes short free-text descriptions of the task, its action space, and its environment state.
Second, following Epoch ECI's naming conventions~\cite{ho2025rosetta}, we normalize model identifiers and scaffold names and standardize release dates. Third, we verify the mapped corpus in two passes. Deterministic tests cover identifier hygiene, date validity, exact-match deduplication on \texttt{(benchmark, task\_id)}, and text quality (HTML entities and control characters). We then conduct manual checks, focusing on newly contributed runs and entries stitched together from multiple sources. Finally, tasks receive SOC occupational~\cite{bls2018soc} and NAICS~\cite{omb2022naics} industry codes following GDPval~\cite{patwardhan2025gdpval} through a three-voter LLM ensemble with 88.3\% (SOC) and 86.8\% (NAICS) majority agreement, with the remainder adjudicated by a stronger model (Appendix~\ref{app:classification}). 

After reconciliation, each record links to its model and scaffold, the environment and task on which it is measured, and the verifier information used in scoring. The pipeline itself is organized as one self-contained builder per source, and the 30 existing builders serve as working examples. Thus, extending the corpus with a new benchmark is a local contribution that leaves the core consolidation logic untouched. We organize the 30 benchmarks into five groups by the type of work the agent performs: programming (software-engineering tasks), research and reasoning (scientific and mathematical problems), enterprise workflows (professional office work), GUI (web and desktop navigation), and function calling (tool selection and invocation). Appendix~\ref{app:benchmark-groups} describes each group in detail.

\begin{table*}[!tp]
  \centering
  \scriptsize
  \setlength{\tabcolsep}{4pt}
  \renewcommand{\arraystretch}{1.0}
  \begin{tabular}{l l l l r r r r}
    \toprule
    \textbf{Benchmark} & \textbf{Benchmark group} & \textbf{Verifier type(s)} & \textbf{Scoring rule}
      & \textbf{\# Tasks} & \textbf{\# Verifiers} & \textbf{\# Records} & \textbf{\# Agents} \\
    \midrule
    \rowcolor{grpProg}     SWE-bench            & Programming          & script      & direct          & 1{,}500 & 1{,}500  & 53\,k   & 92  \\
    \rowcolor{grpProg}     SWE-bench Verified   & Programming          & script      & direct          & 500     & 500      & 129\,k  & 192 \\
    \rowcolor{grpProg}     SWE-bench Pro        & Programming          & script      & direct          & 731     & 731      & 12\,k   & 14  \\
    \rowcolor{grpProg}     GSO                  & Programming          & script      & threshold       & 102     & 102      & 1.5\,k  & 15  \\
    \rowcolor{grpProg}     LiveCodeBench        & Programming          & script      & direct          & 1{,}055 & 1{,}055  & 30\,k   & 22  \\
    \rowcolor{grpProg}     QCircuitBench        & Programming          & script      & threshold       & 28      & 28       & 0.1\,k  & 5   \\
    \rowcolor{grpProg}     TerminalBench        & Programming          & script      & direct          & 89      & 89       & 62\,k   & 146 \\
    \rowcolor{grpProg}     CyBench              & Programming          & exact match & direct          & 15      & 15       & 0.1\,k  & 8   \\
    \addlinespace[2pt]
    \rowcolor{grpResearch} HCAST                & Research \& reasoning & script      & threshold       & 157     & 157      & 15\,k   & 20  \\
    \rowcolor{grpResearch} SWAA                 & Research \& reasoning & script      & direct          & 66      & 66       & 8.3\,k  & 20  \\
    \rowcolor{grpResearch} RE-Bench             & Research \& reasoning & script      & threshold       & 5       & 5        & 0.5\,k  & 20  \\
    \rowcolor{grpResearch} MLE-bench            & Research \& reasoning & script      & threshold       & 82      & 164      & 1.5\,k  & 13  \\
    \rowcolor{grpResearch} ScienceAgentBench    & Research \& reasoning & script      & direct          & 102     & 102      & 0.5\,k  & 5   \\
    \rowcolor{grpResearch} ReplicationBench     & Research \& reasoning & script      & direct          & 90      & 90       & 0.5\,k  & 5   \\
    \rowcolor{grpResearch} HLE                  & Research \& reasoning & judge       & direct          & 1{,}369 & 1{,}369  & 12\,k   & 18  \\
    \rowcolor{grpResearch} MathArena            & Research \& reasoning & exact match, judge & direct, threshold & 352     & 352      & 51\,k   & 84  \\
    \rowcolor{grpResearch} MathHay              & Research \& reasoning & exact match & direct          & 75      & 75       & 1.3\,k  & 5   \\
    \addlinespace[2pt]
    \rowcolor{grpEnt}      TheAgentCompany      & Enterprise workflows & script      & all-pass, threshold & 175     & 541      & 12\,k   & 18  \\
    \rowcolor{grpEnt}      HarveyAI-Lab         & Enterprise workflows & judge       & all-pass        & 1{,}000 & 60{,}187 & 222\,k  & 5   \\
    \rowcolor{grpEnt}      GDPval               & Enterprise workflows & judge       & threshold       & 220     & 220      & 3.5\,k  & 16  \\
    \rowcolor{grpEnt}      MedAgentBench        & Enterprise workflows & script      & direct          & 300     & 300      & 1.5\,k  & 5   \\
    \rowcolor{grpEnt}      $\tau^2$-bench       & Enterprise workflows & exact match, judge, script & all-pass        & 375     & 558      & 25\,k   & 7   \\
    \rowcolor{grpEnt}      MCPBench             & Enterprise workflows & judge       & threshold       & 52      & 312      & 6.3\,k  & 5   \\
    \rowcolor{grpEnt}      DABStep              & Enterprise workflows & script      & direct          & 450     & 450      & 2.2\,k  & 5   \\
    \addlinespace[2pt]
    \rowcolor{grpGui}      OSWorld              & GUI                  & script      & threshold       & 361     & 361      & 31\,k   & 82  \\
    \rowcolor{grpGui}      OnlineMind2Web       & GUI                  & human       & direct          & 300     & 300      & 3.0\,k  & 10  \\
    \rowcolor{grpGui}      BrowseComp           & GUI                  & judge       & direct          & 124     & 124      & 1.8\,k  & 5   \\
    \rowcolor{grpGui}      WebVoyager           & GUI                  & judge       & direct          & 65      & 65       & 1.3\,k  & 5   \\
    \addlinespace[2pt]
    \rowcolor{grpFC}       BFCL-Live            & Function calling     & exact match & direct          & 1{,}351 & 1{,}351  & 147\,k  & 108 \\
    \rowcolor{grpFC}       BFCL-Multi-Turn      & Function calling     & exact match & direct          & 800     & 800      & 87\,k   & 108 \\
    \midrule
    \textbf{Total} & 5 groups & 4 types & 3 rules
      & \textbf{11{,}891} & \textbf{71{,}969} & \textbf{923\,k} & \textbf{745} \\
    \bottomrule
  \end{tabular}
  \caption{\textbf{\messier summary.} Row colors represent benchmark group. Verifier types and scoring rules are defined in \S\ref{sec:preliminaries}. A benchmark may use more than one verifier type, while its scoring rule describes how the final trial result is formed. \emph{Records} counts trial results and their corresponding verifier results. Under direct scoring, the two coincide and are counted once. Records outnumber verifiers because the same verifier may be applied across multiple agents and repeated trials. The Agents column counts unique model--scaffold pairs. Counts are subject to upstream shifts.}
  \label{tab:per-bench}
\end{table*}

\begin{figure*}[!htbp]
  \centering
  \includegraphics[width=\textwidth]{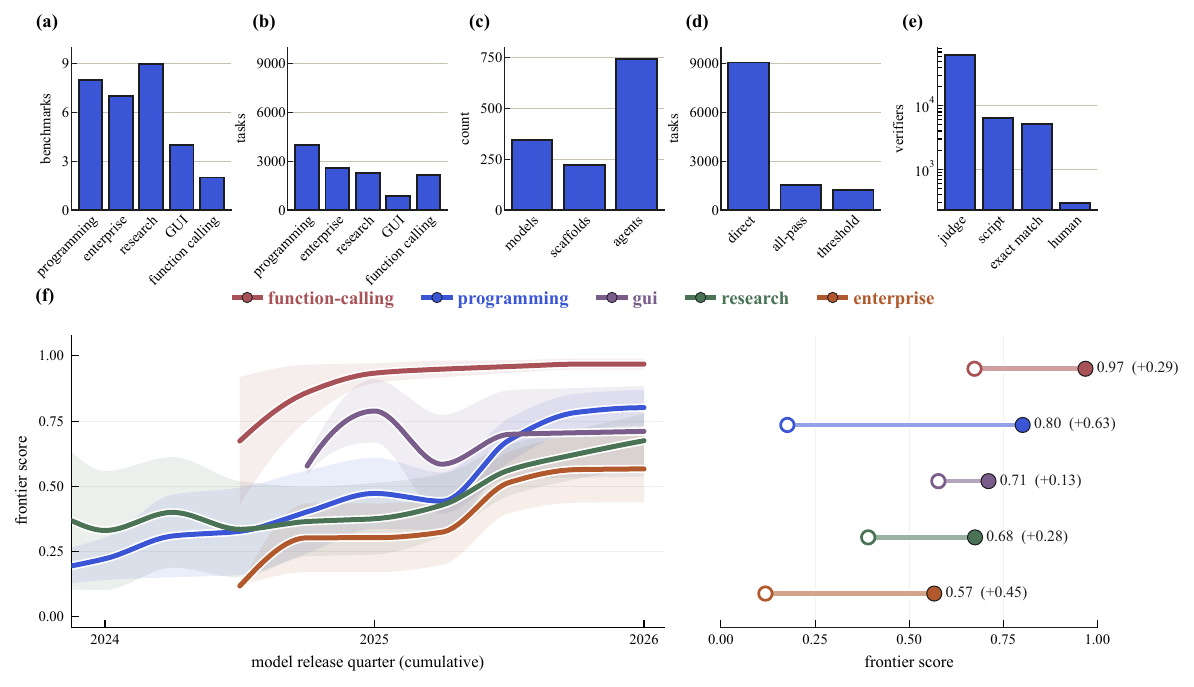}
\caption{\textbf{Composition and frontier analysis.} Panels summarize the corpus across the dimensions defined in \S\ref{sec:preliminaries}:
(a) number of benchmarks by benchmark group,
(b) number of tasks by benchmark group,
(c) models, scaffolds, and agents,
(d) scoring rules frequency,
and (e) verifier type frequency. Panel (e) uses a log scale.
Panel (f) tracks the frontier score by release quarter and benchmark group. Here, the frontier score averages the best result on each task among models released by that quarter. Dumbbell markers connect each group's first and final observed quarter.}
  \label{fig:composition}
\end{figure*}

\section{Analysis \& Applications}
\label{sec:using-messier}

In this section, we present several analyses enabled by \messier, with more in Appendix~\ref{app:additional-analyses}. We begin with a descriptive account of frontier progress across benchmark groups, then show that scoring rules can change measured performance. We demonstrate two applications of the corpus: reproducing capability indices from open data and predicting task difficulty before rollouts.

\subsection{Uneven frontier progress}
\label{sec:analysis-progress}

Figure~\ref{fig:composition} provides a view of frontier progress across the five benchmark groups represented in \messier. 
For each quarter, we consider only agents whose models had been released by then. For each task, we calculate each eligible agent's fraction of successful trials and retain the highest value. The benchmark frontier is the mean of these task values, and the group frontier is the mean of its benchmark frontiers. Because the best agent is selected separately for each task, the resulting frontier need not correspond to the performance of any single agent.
This analysis is not intended to estimate field-wide algorithmic progress, since benchmarks, scaffolds, and evaluation sources differ across groups.

The observed frontier improves across all five benchmark groups between 2024 and 2026, but this progress is asymmetric. Function-calling benchmarks are near saturation with a frontier score of 0.97. Programming follows at 0.80 and shows the largest net growth, improving by 0.63 over the observed window. Enterprise workflows remain the lowest-performing benchmark group at 0.57. This improves over the historical enterprise baseline, but the low absolute score indicates that professional workflows remain difficult for agents. Research and reasoning (0.68) and GUI benchmarks (0.71) occupy the intermediate performance band.

Still, an aggregate score does not, by itself, reveal what agents are failing to do. In a multi-verifier task, an all-pass rule assigns the same trial result of zero to a trial with no positive verifier results and one with nearly all positive verifier results. Final trial results therefore discard the distribution of verifier results. We examine this effect in the subset of \messier benchmarks for which multi-verifier results are available.
\begin{figure*}[t]
  \centering
  \includegraphics[width=\textwidth]{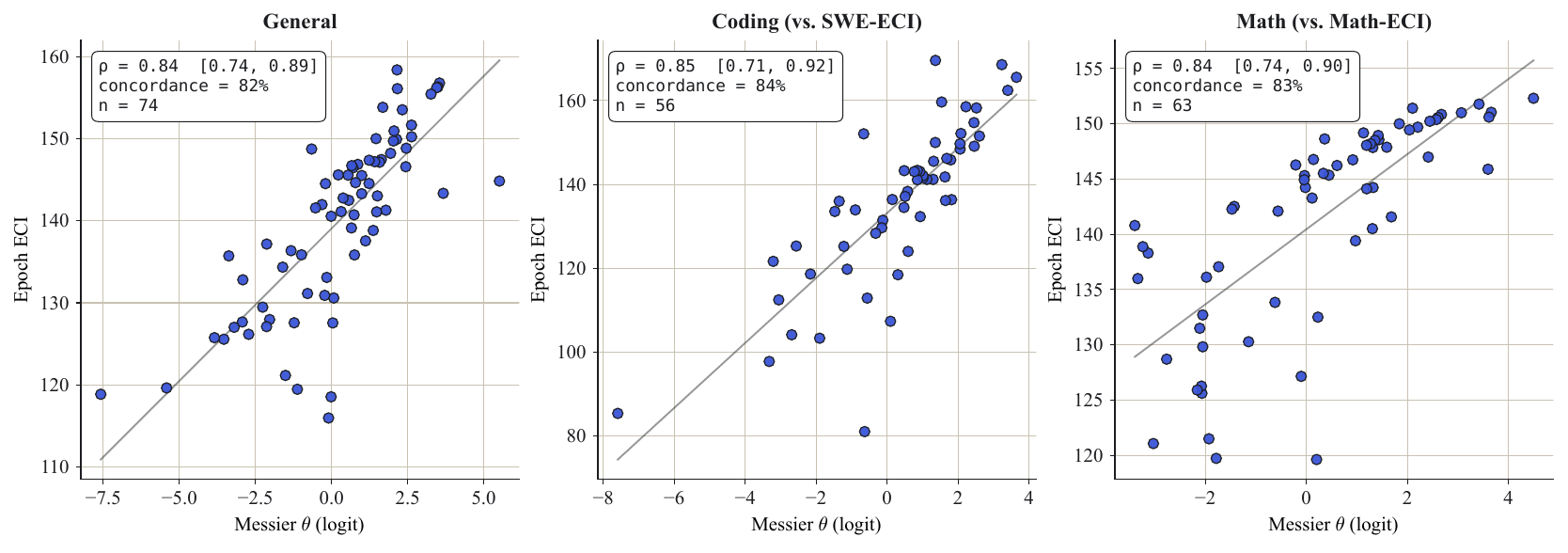}
  \caption{\textbf{Capability orderings from \messier vs.\ Epoch's ECI.} Each panel plots \messier-fit model-level $\theta$ (logit scale) against Epoch's published ECI for the matched subset, with Spearman $\rho$, 95\% bootstrap CI over 1{,}000 resamples, pairwise concordance, and matched-model count. The General panel uses Epoch's overall ECI ($n=74$). Coding uses SWE-ECI ($n=56$). Math uses Math-ECI ($n=63$). A 1PL fit on \messier alone aligns with Epoch's orderings without running new evaluations.}
  \label{fig:eci_reproduction}
\end{figure*}

\subsection{Counterfactual scoring}
\label{sec:aggregation-counterfactual}

The verifier results in \messier let us examine how reported performance changes when the same verifier results are combined differently. We include all three benchmarks in the corpus that provide multiple verifier results per task: HarveyAI-Lab, $\tau^2$-bench, and TheAgentCompany. We rescore the same trials in two counterfactual ways. The first assigns partial credit equal to the fraction of verifiers with a positive result. For example, a trial with three positive results out of four receives a soft score of \(0.75\). The second applies a 50\%-threshold rule, assigning a successful trial result when at least half of its verifier results are positive (denoted $\geq 50\%$). For each benchmark, we report the mean all-pass result, soft score, and 50\%-threshold result across trials. We also report a \emph{near-pass} fraction, measuring how often trials fail all-pass scoring despite having positive results from at least 80\% of their verifiers.

Table~\ref{tab:aggregation-counterfactual} shows that the scoring rule changes measured performance. Across all three benchmarks, the soft and 50\%-threshold scores are substantially higher than the strict all-pass score. The effect is largest on HarveyAI-Lab: only 0.17\% of trials are successful under all-pass scoring, compared with a mean verifier score of 29.27\%. Under 50\%-threshold scoring, 30.28\% of trials are successful.

\begin{table}[t]
  \centering
  \footnotesize
  \setlength{\tabcolsep}{2.5pt}
  \renewcommand{\arraystretch}{1.12}
  \begin{tabular*}{\columnwidth}{@{\extracolsep{\fill}} l r r r r r @{}}
    \toprule
    \textbf{Benchmark}
    & \makecell[c]{\textbf{All-}\\\textbf{pass}}
    & \makecell[c]{\textbf{Soft}\\\textbf{mean}}
    & \makecell[c]{$\mathbf{\geq 50\%}$}
    & \makecell[c]{\textbf{Near-}\\\textbf{pass}}
    & \makecell[c]{$\boldsymbol{\rho}$\\\textbf{(soft,}\\\textbf{all-pass)}} \\
    \midrule
    HarveyAI-Lab        & 0.002 & 0.293 & 0.303 & 0.082 & 0.975 \\
    $\tau^2$-bench      & 0.800 & 0.892 & 0.985 & 0.000 & 1.000 \\
    TheAgentCompany     & 0.151 & 0.340 & 0.352 & 0.011 & 0.986 \\
    \bottomrule
  \end{tabular*}
  \caption{\textbf{Counterfactual scoring on benchmarks with multiple verifier results per task.}
  We report strict all-pass scoring alongside two counterfactual rules: soft (the mean of the binary verifier results), and $\geq 50\%$ (50\%-threshold). Near-pass is the fraction of trials that fail all-pass despite having positive results from at least $80\%$ of their verifiers. The last column reports the Spearman rank correlation across agents between their mean all-pass and mean soft scores. $\rho = 1$ would mean the two rules preserve agent ordering. Lower values indicate the scoring rules reorder agents.}
  \label{tab:aggregation-counterfactual}
\end{table}

The scoring rules substantially change reported performance, but agent orderings remain broadly stable in this sample. Per-agent mean all-pass and mean soft scores correlate at Spearman $\rho = 0.975$ on HarveyAI-Lab, $1.000$ on $\tau^2$-bench, and $0.986$ on TheAgentCompany. These results suggest that evaluators should report verifier results alongside trial results, where available, so that the sensitivity of reported performance to scoring choices remains visible. Further validation across more agents and benchmarks is needed to determine when aggregation also changes comparative conclusions.

\subsection{Open-data capability scales}
\label{sec:eci-reproduction}

We next test whether \messier can recover an established cross-benchmark capability ordering, Epoch's ECI~\cite{ho2025rosetta}, without running new evaluations. To do so, we fit a standard Rasch (1PL IRT) model~\cite{rasch1960probabilistic}. For each model--benchmark pair, we retain the scaffold with the highest fraction of successful trials. 
Under this model, the probability that model $m$ successfully completes task $t$ is
\begin{equation}
    P(Y_{mt} = 1 \mid \theta_m, \beta_t) = \frac{1}{1 + e^{\beta_t - \theta_m}},
\end{equation}
where $Y_{mt} \in \{0, 1\}$ is a trial result for model $m$ on task $t$, $\theta_m$ is the latent model ability, and $\beta_t$ is the task difficulty. Full estimation and optimization details are deferred to Appendix~\ref{app:irt-estimation}.

Since \messier's benchmark composition differs from Epoch's, we assess agreement using Spearman rank correlation ($\rho$) between the resulting model ranking and Epoch's published ECI ranking, rather than comparing absolute values. Moreover, to account for benchmarks that mix multiple skills, we use \messier's task-level classifications to align the comparison by domain. Specifically, we use tasks mapped to SOC 15-1200 for programming and SOC 15-2000 for mathematics.

As shown in Figure~\ref{fig:eci_reproduction}, the resulting 1PL ability estimates ($\theta_m$) closely track the published ECI rankings. We find rank correlations of $\rho = 0.84$ for general capability ($n=74$), $\rho = 0.85$ for programming ($n=56$), and $\rho = 0.84$ for mathematics ($n=63$). A 2PL fit yields nearly identical results on the same data, with Spearman $\rho = 0.99$ relative to the 1PL fit (Appendix~\ref{app:irt-estimation}). Beyond this correspondence with ECI, the design of \messier supports custom capability scales across industries, action spaces, and verifier types.

\subsection{Task difficulty prediction}
\label{sec:difficulty-prediction}

As a final application, we test whether \messier metadata contain signal about task difficulty. Using the fitted IRT task parameters from Section~\ref{sec:eci-reproduction}, we predict two targets: raw difficulty $\beta$ and within-benchmark $z$-scored $\beta$, following the methodology of Agent Psychometrics~\cite{ge2026agent}. We fit a \texttt{RidgeCV} regressor on \texttt{bge-base-en} sentence embeddings~\cite{xiao2024c} using random $k$-fold cross-validation. We compare embeddings of the task instruction alone (\textit{instruction}) with embeddings of the concatenated task instruction, environment description, action-space description, and gold answer (\textit{all}).

Table~\ref{tab:difficulty} shows that the additional metadata yield a modest but consistent improvement across both targets. For raw $\beta$, the model only slightly outperforms the benchmark-mean baseline, suggesting that benchmark identity explains much of the signal. For $z$-scored $\beta$, which removes benchmark-level shifts, the model more clearly outperforms the baseline, indicating that task descriptions carry signal about relative difficulty within benchmarks.

\begin{table}[t]
  \centering
  \small
  \setlength{\tabcolsep}{4pt}
  \renewcommand{\arraystretch}{1.15}
  \begin{tabular}{ll rr}
    \toprule
    \textbf{Target} & \textbf{Input} & \textbf{Model} & \textbf{Baseline} \\
    \midrule
    $\beta$ (raw)
      & instruction & $0.484 \pm 0.008$          & $0.465$ \\
      & all    & $\mathbf{0.508 \pm 0.009}$ & $0.465$ \\
    \addlinespace[2pt]
    $\beta$ ($z$-scored)
      & instruction & $0.263 \pm 0.015$           & $-0.081$ \\
      & all    & $\mathbf{0.298 \pm 0.011}$ & $-0.081$ \\
    \bottomrule
  \end{tabular}
  \caption{\textbf{Difficulty prediction for two targets and input configurations.} Targets: raw IRT difficulty $\beta$ and within-benchmark $z$-scored $\beta$. Inputs: \textit{instruction} = raw instruction text. \textit{all} = concatenation of the instruction, environment, and action-space descriptions plus gold answer. The baseline predicts the training-fold benchmark mean. Entries report Spearman $\rho$ under 5-fold cross-validation, mean $\pm$ SD across folds. $N = 9{,}299$ tasks after filtering.}
  \label{tab:difficulty}
\end{table}

\section{Conclusion}
\label{sec:conclusion}

We introduced \messier, a unified agent evaluation corpus consolidating 30 benchmarks, 745 agents, 11{,}891 tasks, and 71{,}969 verifiers (\S\ref{sec:messier}), including new runs targeting five agents on six benchmarks. By reconciling agents, tasks, environments, verifiers, and scoring rules into a shared schema, \messier makes heterogeneous agent evaluations comparable at both macro- and micro-levels.

Our analyses show that this granularity reveals what aggregate scores can obscure. Frontier performance is uneven across benchmark groups: enterprise workflows remain well below programming and function calling, while programming shows the largest improvement over the observed window. On benchmarks with multiple verifier results per task, counterfactual rescoring shows that all-pass scoring can obscure partial progress and substantially change reported performance. The same records also support capability estimation from open data, aligning with Epoch's ECI ordering (Spearman $\rho = 0.84$), and provide useful signal for difficulty prediction.

Looking ahead, \messier's verifier-result records and execution traces can support studies of scoring, judge design, reward hacking, evaluation awareness, and sandbagging at the granularity where these phenomena occur. Its task classifications and difficulty estimates can also help construct targeted evaluations by occupation, domain, action space, and task difficulty. By reducing the cost of reusing and extending prior evaluations, \messier aims to make future agent evaluations easier to compare, audit, and build upon.

\section*{Limitations}

Several limitations exist. First, the corpus is concentrated in technical, professional, and knowledge-work domains, consists primarily of English-language evaluations, and largely reflects Western software conventions. As a result, findings may not generalize to other languages, cultural settings, or non-technical forms of work. Expanding coverage would require both new benchmarks in underrepresented domains and open per-task releases that make such evaluations available for consolidation.

Second, our collection relies on upstream benchmarks whose construction and quality control are outside our purview. The field is actively pushing the frontier of long-horizon tasks and improving evaluation reliability, as evidenced by the recent emergence of several ``-verified'' benchmark variants. However, this validation remains an ongoing community effort. We therefore inherit the validity assumptions of upstream datasets and do not independently revalidate every task. As these efforts grow, we could track known flawed tasks across benchmarks, which may eventually allow us to model and anticipate these failures.

Third, predicting task difficulty prior to execution remains a significant challenge. Using embeddings for this purpose assumes that difficulty is partly captured by semantic representations, even though embeddings primarily encode semantic content rather than inherent complexity. We made this methodological choice deliberately to align with recent work in this area (e.g., Agent Psychometrics \cite{ge2026agent}). Nevertheless, we believe \textit{a priori} difficulty prediction is a promising direction for future work, with potential benefits in environment design and LLM post-training.

Finally, the counterfactual ranking comparisons contain relatively few agents per benchmark. Their correlations should therefore be interpreted as descriptions of the observed runs rather than general estimates of how aggregation affects agent rankings.

\section*{Ethical Considerations}

As task horizons lengthen \cite{kwa2026measuring}, the financial and environmental costs of running evaluations grow substantially \cite{strubell2019energy}, widening the gap between well-resourced and under-resourced groups. While \messier is designed to alleviate this burden by minimizing redundant evaluation runs, the underlying compute requirements remain a structural equity concern. To partially bridge this gap, we have spent approximately \$10{,}000 on the contributed evaluations. By standardizing and distributing these evaluation results publicly, we hope to lower the barrier to entry for researchers who lack the capital to run large-scale evaluations.

Furthermore, agent benchmarks frequently abstract or simulate work performed by paid human workers. Reading quantitative scores in isolation risks obscuring the socioeconomic implications for the workers whose tasks are being automated. Evaluation releases must therefore maintain a clear accounting of whose labor is being measured and whom that measurement serves.

Finally, our release preserves upstream license and access conditions as described in Appendix~\ref{app:licensing}, and it does not relicense third-party benchmark materials.

\section*{Acknowledgments}
We thank Clément Tisseau for identifying an issue in the handling of failed HarveyAI-Lab trials and for providing reproduction code that helped us correct the data and analysis.

\bibliography{custom}
\clearpage

\appendix

\section{Dataset Details}

\subsection{Benchmark groups}
\label{app:benchmark-groups}

\paragraph{Programming.}
Programming benchmarks place the agent in a software-engineering environment, typically a source repository or terminal session, and evaluate whether the resulting code or command output satisfies executable checks. The group includes SWE-bench, SWE-bench Verified, and SWE-bench Pro, along with LiveCodeBench and GSO for code generation against test suites, TerminalBench and CyBench for shell-level engineering and capture-the-flag tasks, and QCircuitBench for quantum algorithm design. Although SWE-bench Verified is a curated subset of SWE-bench upstream, we treat it as a separate benchmark.

\paragraph{Research and reasoning.}
Research and reasoning benchmarks ask the agent to solve a scientific, mathematical, or research-engineering problem and to produce either a correct answer or a successful experimental artifact. The group splits between research benchmarks that produce an artifact or analytical solution (HCAST, SWAA, RE-Bench, MLE-bench, ScienceAgentBench, ReplicationBench), and reasoning benchmarks that produce an answer to a difficult problem in mathematics or general knowledge (HLE, MathArena, MathHay). HCAST, SWAA, and RE-Bench additionally provide per-task human completion-time bands used in our calibration analysis.

\paragraph{Enterprise workflows.}
Enterprise workflow benchmarks place the agent in delegated professional or office work, including legal review, clinical workflows, customer support, financial analysis, and general knowledge-work tasks. The group includes HarveyAI-Lab, GDPval, MCPBench, TheAgentCompany, MedAgentBench, DABStep, and $\tau^2$-bench. Several provide multiple verifier results per task, making it possible to study partial completion even when the final trial result is binary.

\paragraph{GUI navigation.}
GUI benchmarks evaluate the ability of the agent to operate in graphical or web environments through clicks, keystrokes, and page navigation, with success measured by the final state of the environment. The group includes OSWorld for desktop applications, OnlineMind2Web for live-web navigation, BrowseComp for web research, and WebVoyager for browsing tasks.

\paragraph{Function calling.}
Function-calling benchmarks evaluate whether agents can select and invoke the appropriate tools, often from large tool sets with overlapping function names, required inputs, and contextual constraints such as user roles or permissions. The group consists of BFCL-Live for single-turn function calls and BFCL-Multi-Turn for multi-turn conversations over stateful tools.

\subsection{Sources}
\label{app:sources}
Table~\ref{tab:per-bench-sources} lists the upstream source for each of the 30 \messier benchmarks. \emph{HF} denotes a Hugging Face dataset path. \emph{GitHub} denotes a public repository (pinned commits and dataset revisions are recorded). For benchmarks consolidated from secondary releases (BRIDGE, Agent Psychometrics, General AgentBench), we record the consolidator alongside the upstream owner. Harbor-collected entries are those we evaluated ourselves under the OpenHands scaffold (\S\ref{sec:messier}).

\subsection{SOC and NAICS classification}
\label{app:classification}
Tasks receive SOC occupational and NAICS industry codes through a three-tier pipeline. First, three voter models (Anthropic Claude Haiku, Anthropic Claude Sonnet, and OpenAI GPT-5-mini) independently classify each task against the full BLS 2018 SOC structure and Census 2022 NAICS sector list, prompted with the task description and a brief gloss of every code. Second, splits among the three voters are adjudicated by Anthropic Claude Opus on the disputed subset. Third, rule-based keyword overrides handle a small set of edge cases. Benchmarks whose tasks deterministically map to a single occupation (e.g., the SWE-bench family to Computer Occupations) bypass the voter pipeline and are crosswalked directly. Across the 4{,}311 tasks rated by all three voters, the voters reach 2-of-3 agreement on 88.3\% of SOC labels and 86.8\% of NAICS labels. Cases without majority agreement are adjudicated by Opus.

\subsection{Data model and terminology}
\label{app:model}
Table~\ref{tab:terminology} summarizes the terminology used throughout the paper, and Figure~\ref{fig:schema-er} shows how the entities relate.

\begin{table*}[h]
\centering
\small
\begin{tabular}{p{0.18\textwidth} p{0.76\textwidth}}
\toprule
\textbf{Term} & \textbf{Definition} \\
\midrule

Model
& The underlying language model. \\

\cellcolor{grpP}Scaffold
& \cellcolor{grpP}The system that turns model outputs into actions and parses environment observations. \\

Agent
& A model--scaffold pair. \\

\cellcolor{grpP}Environment
& \cellcolor{grpP}The system with which an agent interacts, defining the states, actions, and observations available. \\

Task
& A given problem consisting of an instruction and initial environment state. \\

\cellcolor{grpP}Trial / rollout
& \cellcolor{grpP}One execution of one agent on one task. \\

Verifier
& A procedure that evaluates a trial, such as a script, LLM judge, exact-match check, or human evaluation. A verifier may internally perform multiple checks. \\

\cellcolor{grpP}Verifier result
& \cellcolor{grpP}The binary or numerical value returned by one verifier for one trial. \\

Scoring rule
& The rule that maps one or more verifier results to a trial result (typically defined per benchmark). \\

\cellcolor{grpP}Trial result
& \cellcolor{grpP}The value $S \in \{0,1\}$ assigned to a trial by the scoring rule. \\

Record
& A stored row in \messier containing a trial result or verifier result. For example, one trial with three verifier results may produce four records: three verifier-result records and one trial-result record. Some records contain only source-reported summaries such as pass@k (3.1\% in total). \\

\bottomrule
\end{tabular}
\caption{Terminology used throughout the paper.}
\label{tab:terminology}
\end{table*}

\subsection{Contributed runs configuration}
\label{app:compute}
The six contributed benchmarks (HarveyAI-Lab, MedAgentBench, DABStep, QCircuitBench, ScienceAgentBench, and ReplicationBench) were collected under a single fixed configuration through the Harbor harness against hosted model APIs. We evaluated five agents, each pairing one of the following models with the same scaffold: \nolinkurl{openai/gpt-5}, \nolinkurl{openai/gpt-5-nano}, \nolinkurl{openai/gpt-4o}, \nolinkurl{anthropic/claude-haiku-4-5}, and \nolinkurl{anthropic/claude-sonnet-4-5}. Each (agent, task) pair was run once. The shared scaffold is \nolinkurl{openhands-sdk} v1.22.0, executed inside Modal sandbox images. HarveyAI-Lab uses 23--194 verifier results per task, each produced by the LLM judge \nolinkurl{anthropic/claude-sonnet-4-6}. Under its all-pass scoring rule, every verifier result must equal $1$ for the trial result to be $S=1$. The total spend was approximately \$10{,}000.

\subsection{Licensing}
\label{app:licensing}

\messier is a mixed-license collection assembled from third-party benchmarks with heterogeneous licenses and access conditions. We do not claim that third-party benchmark materials are relicensed under a single umbrella license. Each upstream component remains subject to its original terms, and Table~\ref{tab:per-bench-sources} lists the public source and the corresponding license or access status we identified at the time of release.

Our original contributions in this work, including normalization code, schema definitions, benchmark mappings, and released derived artifacts such as summary statistics or result tables, are licensed under the MIT License. This license applies only to our original contributions and does not override or replace any upstream license.

When an upstream source is gated, lacks a clear public license, or presents ambiguous redistribution terms, we do not redistribute the underlying task content. Instead, where permitted, we release only minimal derived artifacts such as normalized identifiers and/or pass/fail results, and direct users to obtain the original data from the upstream provider.

\begin{table*}[!htbp]
  \centering
  \scriptsize
  \setlength{\tabcolsep}{4pt}
  \renewcommand{\arraystretch}{1.0}
  \begin{tabular}{@{}l l l@{}}
    \toprule
    \textbf{Benchmark} & \textbf{Source} & \textbf{Upstream license / access status} \\
    \midrule
    SWE-bench            & GitHub: \texttt{princeton-nlp/SWE-bench} & MIT \\
    SWE-bench Verified   & \makecell[l]{HF: \texttt{princeton-nlp/SWE-bench\_Verified} \\ aux.\ GitHub: \texttt{dariakryvosheieva/agent-psychometrics} \\ aux.\ GitHub: \texttt{McGill-NLP/BRIDGE} (human-time)} & \makecell[l]{MIT via SWE-bench repo \\ HF tag not stated} \\
    SWE-bench Pro        & \makecell[l]{HF: \texttt{ScaleAI/SWE-bench\_Pro} \\ aux.\ GitHub: \texttt{dariakryvosheieva/agent-psychometrics}} & no explicit standard license identified \\
    GSO                  & \makecell[l]{HF: \texttt{gso-bench/gso} \\ aux.\ GitHub: \texttt{dariakryvosheieva/agent-psychometrics}} & MIT \\
    LiveCodeBench        & HF: \texttt{livecodebench/code\_generation\_lite} & CC (HF tag, variant unspecified) \\
    QCircuitBench        & Harbor (this work) & MIT \\
    TerminalBench        & \makecell[l]{GitHub: \texttt{laude-institute/terminal-bench} \\ via GitHub: \texttt{dariakryvosheieva/agent-psychometrics}} & Apache-2.0 \\
    CyBench              & \makecell[l]{GitHub: \texttt{andyzorigin/cyber-bench} \\ via GitHub: \texttt{McGill-NLP/BRIDGE} (results + human-time)} & Apache-2.0 \\
    \midrule
    HCAST                & \makecell[l]{GitHub: \texttt{METR/hcast-public} \\ runs via GitHub: \texttt{METR/eval-analysis-public}} & MIT$^\dagger$ \\
    SWAA                 & runs via GitHub: \texttt{METR/eval-analysis-public} & no explicit standard license identified \\
    RE-Bench             & \makecell[l]{GitHub: \texttt{METR/RE-Bench} \\ runs via GitHub: \texttt{METR/eval-analysis-public}} & MIT$^\dagger$ \\
    MLE-bench            & \makecell[l]{Kaggle competitions \\ via GitHub: \texttt{McGill-NLP/BRIDGE}} & \makecell[l]{MIT for code \\ competition data subject to Kaggle terms} \\
    ScienceAgentBench    & Harbor (this work) & MIT \\
    ReplicationBench     & Harbor (this work) & MIT \\
    HLE                  & \makecell[l]{HF: \texttt{cais/hle} \\ aux.\ GitHub: \texttt{supaihq/hle} (judged-pro labels)} & MIT, gated, results only$^\ddagger$ \\
    MathArena            & HF: \texttt{MathArena/\{slug\}\_outputs} & CC-BY-NC-SA-4.0 \\
    MathHay              & HF: \texttt{cx-cmu/agent\_trajectories} & gated, no explicit HF license identified$^\S$ \\
    \midrule
    TheAgentCompany      & \makecell[l]{GitHub: \texttt{TheAgentCompany/TheAgentCompany} \\ GitHub: \texttt{TheAgentCompany/experiments}} & \makecell[l]{MIT for benchmark repo \\ no explicit license identified for experiments repo} \\
    HarveyAI-Lab         & Harbor (this work) & no explicit standard license identified \\
    GDPval               & \makecell[l]{HF: \texttt{openai/gdpval} \\ aux.\ GitHub: \texttt{McGill-NLP/BRIDGE}} & no explicit standard license identified \\
    MedAgentBench        & Harbor (this work) & MIT \\
    $\tau^2$-bench       & \makecell[l]{GitHub: \texttt{sierra-research/tau2-bench} \\ submissions: Sierra S3} & MIT \\
    MCPBench             & HF: \texttt{cx-cmu/agent\_trajectories} & gated, no explicit HF license identified$^\S$ \\
    DABStep              & Harbor (this work) & CC-BY-4.0 \\
    \midrule
    OSWorld              & \makecell[l]{GitHub: \texttt{xlang-ai/OSWorld} \\ HF: \texttt{xlangai/ubuntu\_osworld\_verified\_trajs}} & \makecell[l]{Apache-2.0 for code \\ MIT for HF trajectories} \\
    OnlineMind2Web       & \makecell[l]{HF: \texttt{osunlp/Online-Mind2Web} \\ labels: \texttt{osunlp/Online\_Mind2Web\_Leaderboard} (HF Spaces)} & CC-BY-4.0, gated \\
    BrowseComp           & HF: \texttt{cx-cmu/agent\_trajectories} & gated, no explicit HF license identified$^\S$ \\
    WebVoyager           & HF: \texttt{cx-cmu/agent\_trajectories} & gated, no explicit HF license identified$^\S$ \\
    \midrule
    BFCL-Live            & GitHub: \texttt{ShishirPatil/gorilla} (\texttt{berkeley-function-call-leaderboard}) & Apache-2.0 \\
    BFCL-Multi-Turn      & GitHub: \texttt{ShishirPatil/gorilla} (\texttt{berkeley-function-call-leaderboard}) & Apache-2.0 \\
    \bottomrule
  \end{tabular}
  \caption{Per-benchmark sources and upstream license/access status. Group ordering follows Table~\ref{tab:per-bench}. \emph{aux.} marks secondary releases used to cross-reference or fill per-task gaps under normalized identifiers. The license/status column reports the terms identified for the cited upstream source. It does not define the license of \messier as a whole. Our original \messier code, schema, mappings, and derived artifacts are released under MIT, without overriding upstream licenses or access conditions. $^\dagger$MIT with author requests to avoid publishing unprotected solutions or including tasks in LLM training data. $^\ddagger$HLE is tagged MIT on Hugging Face but is gated and asks users not to publicly share, re-upload, or distribute the dataset. We therefore redistribute only derived pass/fail results. $^\S$The cited Hugging Face trajectory release is gated and no explicit HF license tag was identified. Underlying benchmark projects may have separate licenses, but we do not treat those as relicensing the trajectory release.}
  \label{tab:per-bench-sources}
\end{table*}

\section{Methodology}
\label{app:methodology}

\subsection{Human-time calibration}
\label{app:bridge}
Under BRIDGE's exact methodology (METR-band tasks, 2PL IRT), \messier's pipeline aligns with their log-linear slope within the bootstrap CI: slope $= 0.80$, 95\% CI $[0.71, 0.90]$, $n = 158$. The corresponding doubling factor per unit of latent capability is 2.23, matching BRIDGE's published 2.26 and METR's 2.36. We report this as an implementation check confirming pipeline consistency on overlapping data.

\subsection{Item response theory}
\label{app:irt-estimation}
We fit the 1PL Rasch model $P(Y_{mt}=1) = \sigma(\theta_m - \beta_t)$ via stochastic variational inference in Pyro, with each available trial result included separately. We use normal priors on $\theta_m$ and $\beta_t$ (mean $0$, standard deviation $1$), mean-field normal guides, and center $\theta_m$ to have mean zero during fitting for identifiability. Optimization uses \texttt{ClippedAdam} with learning rate $0.01$ over 2{,}000 epochs (seed $42$). For the BRIDGE-aligned replication (Appendix~\ref{app:bridge}) we additionally fit a 2PL variant, $P(Y_{mt}=1) = \sigma(\alpha_t(\theta_m - \beta_t))$, with a log-normal discrimination prior on $\alpha_t$. On the full corpus, the 1PL and 2PL model orderings agree at Spearman $\rho = 0.99$, so we report 1PL throughout. Tasks for which every retained trial has the same result (all successful or all unsuccessful) are dropped before fitting. The resulting $\theta$ estimates are only comparable when models have meaningful overlapping benchmark coverage, and estimates for sparsely observed models should not be interpreted as global rankings. Domain-restricted ability estimates (e.g. for SOC 15-1200 coding tasks) are obtained by holding $\beta_t$ fixed at its full-corpus posterior mean and refitting $\theta_m$ on the domain subset only, which preserves cross-domain comparability.



\subsection{IRT parameter distributions}

\begin{figure}[h]
    \centering
    \includegraphics[width=1.0\linewidth]{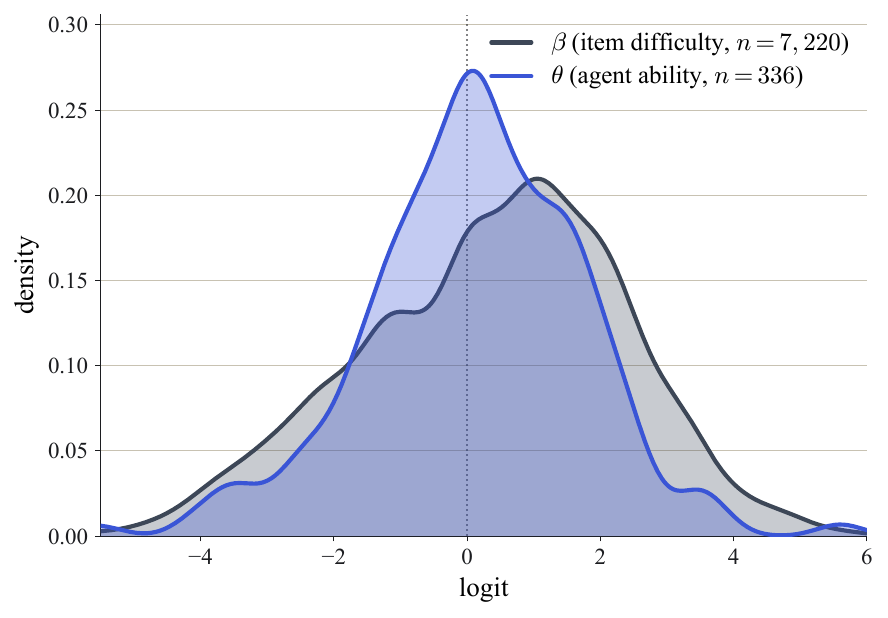}
    \caption{Distributions of fitted 1PL IRT model ability $\theta$ and task difficulty $\beta$ across the full \messier corpus.}
    \label{fig:irt-distributions}
\end{figure}

The distributions of the 1PL fit used throughout the paper are shown in Figure~\ref{fig:irt-distributions}. Both are broadly bell-shaped, although model ability is left-skewed (skewness $-0.98$, mean $0$ by construction), while task difficulty is closer to symmetric (skewness $-0.27$, mean $0.38$). Thus, fitted task difficulty is slightly higher on average than centered model ability, although the distributions overlap substantially.

\label{app:irt-distributions}

\section{Additional Analyses}
\label{app:additional-analyses}

\paragraph{Benchmark groups.}
To examine model capability across \messier, we fit separate 1PL IRT models for each of the five benchmark groups: programming, research and reasoning, enterprise workflows, GUI, and function calling. For each agent and benchmark, we calculate the fraction of successful trials on each task. We then average these fractions so every task contributes equally to the benchmark score. This prevents tasks with more repeated trials from receiving greater weight. For each model--benchmark pair, we retain the agent with the highest benchmark score. Within each benchmark group, we retain models with at least 30 trial results and exclude tasks whose retained trial results are all identical. We fit a group only when at least 20 eligible models remain. All five groups meet this requirement. We standardize the ability estimates ($\theta$) within each group and construct a model-by-group matrix, retaining models represented in at least two groups. We replace missing entries with zero, corresponding to the mean standardized ability within that group. The resulting matrix contains 55 models and all five benchmark groups.

We apply principal component analysis (PCA) to this matrix. The first principal component explains 54.8\% of the variance across models, while three are required to explain 80\%. This indicates a substantial common performance axis alongside meaningful variation across benchmark groups.

\paragraph{Cross-benchmark dependence.}
We examine whether the benchmarks capture distinct patterns of model performance or largely reproduce the same model ranking. Unlike the group-level analysis above, this analysis operates at the finer-grained benchmark level. For each agent and benchmark, we calculate the fraction of successful trials on each task. We then average these fractions so every task contributes equally to the benchmark score. We exclude agents evaluated on fewer than three tasks and, for each model--benchmark pair, retain the agent with the highest benchmark score.

Because of sparsity, we repeat the analysis using four coverage requirements, retaining benchmarks evaluated with at least 5, 10, 15, or 20 models and models represented in at least two retained benchmarks. We reapply both requirements after filtering. This produces matrices containing 169 models and 28 benchmarks, 166 models and 14 benchmarks, 162 models and 9 benchmarks, and 162 models and 9 benchmarks, respectively.

To focus on relative model performance, we rank the models separately within each benchmark and standardize these ranks to zero mean and unit variance. When a model has no result for a retained benchmark, we assign it the mean standardized rank for that benchmark. We then apply PCA to the resulting model-by-benchmark matrix. If the benchmarks produced essentially the same model ordering, a single principal component would explain most of the variation. Instead, across the 5-, 10-, 15-, and 20-model coverage thresholds, the first component explains only 25.0--33.5\% of the variance, while eight, six, four, and four components, respectively, are required to explain 80\%. The benchmarks, therefore, share a common performance signal but also retain substantial benchmark-specific variation. Their model rankings are thus not reducible to a single common ordering.

\end{document}